\RequirePackage{fix-cm}

\documentclass[twocolumn]{svjour3}   

\smartqed
\RequirePackage{fix-cm}
\usepackage{graphicx}
\usepackage{subfigure} 
\usepackage{algorithm}
\usepackage{algorithmic}
\usepackage{amsmath}
\usepackage{dsfont}
\usepackage{breakcites}
\usepackage{subfigure}
\usepackage{lineno,hyperref}
\modulolinenumbers[5]

\begin{document}

\title {Network of Bandits insure Privacy of end-users}

\author{Rapha\"{e}l F\'{e}raud}

\institute{Rapha\"{e}l F\'{e}raud  \at
              Orange Labs, 
       				2, avenue Pierre Marzin,
       				22307, Lannion, France\\
              \email{raphael.feraud@orange.com}  
}

\date{Received: date / Accepted: date}

\maketitle

\begin{abstract}
In order to distribute the best arm identification task as close as possible to the user's devices, 
on the edge of the Radio Access Network, we propose a new problem setting, where distributed players collaborate to find the best arm.
This architecture guarantees privacy to end-users since no events are stored. The only thing that can be observed by an adversary through the core network is aggregated information across users. We provide a first algorithm, {\sc Distributed Median Elimination}, which is optimal in term of number of transmitted bits and near optimal in term of speed-up factor with respect to an optimal algorithm run independently on each player.
In practice, this first algorithm cannot handle the trade-off between the communication cost and the speed-up factor, and
requires some knowledge about the distribution of players.
{\sc Extended Distributed Median Elimination} overcomes these limitations, by playing in parallel different instances of {\sc Distributed Median Elimination} and selecting the best one.
Experiments illustrate and complete the analysis. According to the analysis, in comparison to {\sc Median Elimination} performed on each player,
the proposed algorithm shows significant practical improvements.

\keywords{privacy, distributed algorithm, multi-armed bandits, best arm identification, PAC learning, sample complexity.}
\end{abstract}

\section {Introduction}
\subsection{Motivation}

\paragraph {}
Big data systems store billions of events generated by end-users. 
Machine learning algorithms are then used for instance to infer intelligent mobile phone applications, 
to recommend products and services, to optimize the choice of ads, to choose the best human machine interface, to insure self-care of set top boxes...
In this context of massive storage and massive usage of models inferred from personal data, privacy is an issue. 
Even if individual data are anonymized, the pattern of data associated with an individual is itself uniquely identifying.
The $k$-anonymity approach \cite {S02} provides a guarantee to resist to direct linkage between stored data and the individuals.
However, this approach can be vulnerable to composition attacks: an adversary could use side information that combined with the $k$-anonymized data allows
to retrieve a unique identifier \cite {GKS08}.
Differential privacy \cite {SC13} provides an alternative approach. The sensitive data are hidden. 
The guarantee is provided by algorithms that allow to extract information from data. 
An algorithm is differentially private if the participation of any record in the database does not alter the probability of any outcome by very much.
The flaw of this approach is that, sooner or later, the sensitive data may be hacked by an adversary. Here, we propose to use a radical approach to insure privacy, 
that is a narrow interpretation of {\it privacy by design}. 

\paragraph{}
Firstly, the useful information is inferred from the stream without storing data. 
As in the case of {\it differential privacy}, this {\it privacy by design} approach needs specific algorithms to infer useful information from the data stream.
A lot of algorithms has been developed for stream mining, and most business needs can be handled without storing data: basic queries and statistics can be done on the data stream \cite {BBDMW02}, 
as well as queries on the join of data streams \cite {CMN99,FCG09}, online classification \cite {DH02}, online clustering \cite {BH06}, 
and the more challenging task of decision making using contextual bandits \cite {CLRS11,FAUC16}.
However, even if the data are not stored, the guarantee is not full: an adversary could intercept the data, then stores and deciphers it.

Secondly, to make the interception of data as expensive as possible for an adversary, we propose to locally process the personal data, benefiting from a new network architecture.
For increasing the responsiveness of mobile phone services and applications, network equipment vendors and mobile operators specified a new network architecture: Mobile Edge Computing (MEC) provides IT and cloud computing capabilities within the Radio Access Network in close proximity to devices \cite {MEC14}.
In addition to facilitate the distribution of interactive services and applications, the distribution of machine learning algorithms on MEC 
makes the interception task more difficult. As the data are locally processed, the adversary has to locally deploy and maintain technical devices or software to intercept and decipher the radio communication between devices and MEC servers.

\subsection {Related works}

\paragraph {}
Most of applications necessitate to take and optimize decisions with a partial feedback. 
That is why this paper focuses on a basic block which is called {\it multi-armed bandits} ({\sc mab}). 
In its most basic formulation, it can be stated as follows:
there are $K$ arms, each having an unknown distribution of bounded rewards. At each step, the player has to choose an arm and receives a reward.
The player needs to explore to find profitable arms, but on other hand the player would like to exploit the best arms as soon as possible: this is the so-called exploration-exploitation dilemna.
The performance of a {\sc mab} algorithm is assessed in term of {\it regret} (or opportunity loss) with regards to the unknown optimal arm.
Optimal solutions have been proposed to solve this problem  using a stochastic formulation in \cite{ABF02,CGMMS13}, using a Bayesian formulation in \cite {KKM12}, 
or using an adversarial formulation in \cite {ABFS02}. 

\paragraph {}
The best arm identification task consists in finding the best arm with high probability while minimizing the number of times suboptimal arms are sampled, which corresponds to minimize the regret of the exploitation phase while minimizing the cost of the exploration phase.
While the regret minimization task has its roots in medical trials, where it is not acceptable to give a wrong treatment to a sick patient for exploration purpose, 
the best arm identification has its roots in pharmaceutical trials, where in a test phase the side effects of different drugs are explored, 
and then in a exploitation phase the best drug is produced and sold. The same distinction exists for digital applications, where for instance the regret minimization task is used for ad-serving, 
and the best arm identification task is used to choose the best human machine interface. 
Corresponding to these two related tasks, the fully sequential algorithms, such as {\sc UCB} \cite{ABF02}, explore and exploit at the same time, while the explore-then-commit algorithms, such as {\sc Successive Elimination} \cite {EMM02}, consist in exploring first to eliminate sequentially the suboptimal arms thanks to a statistical test, and then in exploiting the best arm (see \cite {PRCS16} for a formal description of explore-then-commit algorithms).
The analysis of explore-then-commit algorithms is based on the PAC setting \cite {V84}, and focuses on the sample complexity (i.e. the number of time steps) needed to find an $\epsilon$-approximation of the best arm with a failure probability $\delta$. This formulation has been studied for best arm identification problem in \cite {EMM02,BMS09,ABM10,GGL13}, for dueling bandit problem in \cite {UCFN13}, for linear bandit problem in \cite {SLM14}, for the contextual bandit problem in \cite {FAUC16}, and for the non-stationary bandit problem in \cite {RFM17}.

\paragraph {}
Recent years have seen an increasing interest for the study of the collaborative distribution scheme: $N$ players collaborate to solve a multi-armed bandit problem. 
The distribution of non-stochastic experts has been studied in \cite {KLR12}. 
The distribution of stochastic multi-armed bandits has been studied for peer to peer network in \cite {SBHOJK13}. 
In \cite {HKKLS13}, the analysis of the distributed exploration is based on the sample complexity need to find the best arm with an approximation factor $\epsilon$. 
When only one communication round is allowed, an algorithm with an optimal speed-up factor of $\sqrt{N}$ has been proposed. The algorithmic approach has been extended to the case where multiple communication rounds are allowed. In this case a speed-up factor of $N$ is obtained while the number of communication rounds is in $O(\ln1/\epsilon)$. The authors focused on the trade-off between the number of communication rounds and the number of pulls per player. This analysis is natural when one would like to distribute the best arm identification task on a centralized processing architecture. In this case,
the best arm identification tasks are synchronized and the number of communication rounds is the true cost. 

\paragraph {}
The distribution of bandit algorithms on MEC, that we would like to address, is more challenging. 
When bandit algorithms are deployed close to the user's devices, the event {\it player is active} is modeled by an indicator random variable. 
Indeed, a player can choose an action only when an uncontrolled event occurs such as: the device of a user is switched on, a user has launched a mobile phone application, a user connects to a web page...  Unlike in \cite {HKKLS13}, where the draw of players is controlled by the algorithm, here we consider that the players are drawn from a distribution.  As a consequence, synchronized communication rounds can no longer be used to control the communication cost. Here the cost of communications is modeled by the number of transmitted bits. 

\subsection {Our contribution}

\paragraph {}
Between the two main formulations of bandit algorithms, the regret minimization and the best arm identification tasks, we have chosen to distribute the best arm identification task for two reasons.
Firstly, even if it has been shown that the explore-then-commit algorithms are suboptimal for the regret minimization task with two arms by a factor $2$ \cite {GKL16}, 
they can be rate optimal for the regret minimization task, while the fully sequential algorithms cannot handle the best arm identification task.
By distributing an explore-then-commit algorithm, one can provide a reasonably good solution for the two tasks.
Secondly, for distributing bandit algorithms, explore-then-commit algorithms have a valuable property: the communications between players are needed only during the exploration phase.
For each distributed best arm identification task, one can bound the communication cost and the time interval where communications are needed.
This property facilitates the sharing of the bandwith between several distributed tasks.

\paragraph {}
In the next section, we propose a new problem setting handling the distribution of the best arm identification task between collaborative players.
A lower bound states the minimum number of transmitted bits needed to reach the optimal speed-up factor $\mathcal{O}(N)$.
Then, we propose a first algorithm, {\sc Distributed Median Elimination}, which is optimal in term of number of transmitted bits, 
and which benefits from a near optimal speed-up factor with respect to a rate optimal algorithm such as {\sc Median Elimination} \cite {EMM02} run on a single player.
This first algorithm is designed to obtain an optimal communication cost. In practice, it cannot handle the trade-off between the communication cost and the exploration cost, ant it requires
some knowledge on the distribution of players.
{\sc Extended Distributed Median Elimination} overcomes these limitations, by playing in parallel different instances of {\sc Distributed Median Elimination} and selecting the best one.
In the last section, experiments illustrate the analysis of proposed algorithms.

\section {Problem setting}

\begin {figure}[ht]
\begin{center}
  {\includegraphics[width=8.5cm]{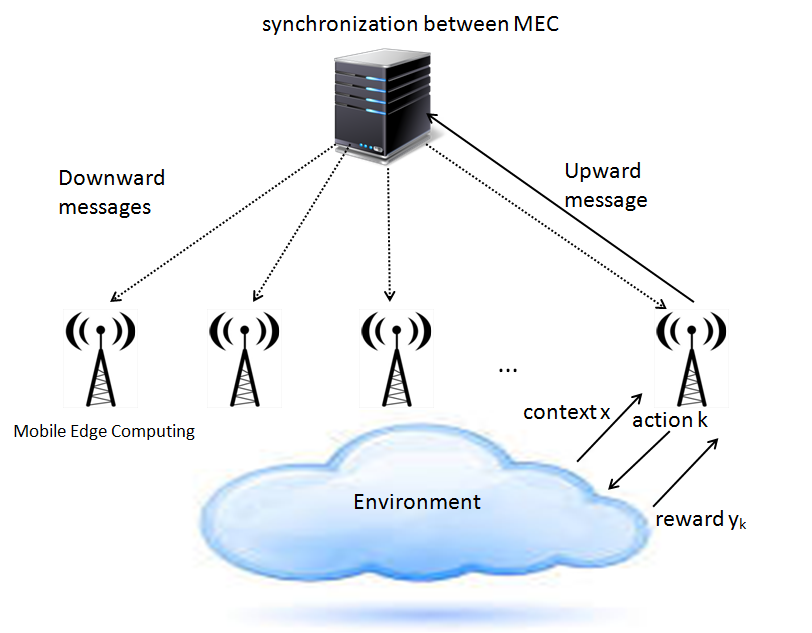}}
\end{center}
\vspace {-0.55cm}
\caption{Principle: the events are processed on the Mobile Edge Computing (MEC) application servers, and the synchronization server shares information between MEC.}
\label{fig1}
\end {figure}

\paragraph {}
The distribution of the best arm identification task on the edge of the Radio Access Network is a collaborative game, where the players are the Mobile Edge Computing application servers, which cluster end-users.
The players attempt to find the best arm as quickly as possible while minimizing the number of sent messages. There are two kinds of messages: the upward messages are sent from the MEC to the synchronization server, and the downward messages are sent from the synchronization server to all MEC (see Figure \ref {fig1}). This architecture handles the case where a context is observed before the action is chosen. The context can contain aggregated information at the MEC level or personal data stored beforehand in the device of the end-user.  
In the following, we focus on the case where no context is observed. We discuss the extension of the proposed algorithm to the {\it contextual bandit} in the future works. 
This architecture guarantees privacy since no event are stored. The part of the context containing personal data, which can be declarative data provided with an opt-in, are under the control of the end-user: if a context is stored in the user's device, it can be suppressed by the end-user. Furthermore, the context can be built in order to insure $k$-anonymity or differential privacy.
The only thing that can be observed by an adversary through the core network (between the MEC and the synchronization server) is upward messages, 
which corresponds to aggregated information across users of one MEC server, and downward messages, which correspond to aggregated information by all MEC servers.
As personal data are locally processed, the adversary has to locally deploy and maintain technical devices or software to intercept and decipher the radio communication between devices and MEC servers. For the adversary, this makes expensive the data collection task.

\paragraph {}
Let $\mathcal {N}$ be the set of players, and $N$ be the number of players. Let $n$ be a random variable denoting the {\it active} player (i.e. the player for which an event occurs), and $P(n)$ be the probability distribution of $n$.
Let $\gamma \in [0,1)$ and $\mathcal {N}_\gamma = \{i \leq N, P(n=i) \geq \gamma \}$ be the set of indices of most active players, and $N_\gamma$ be the number of most active players.
Let $\mathcal{K}$ be a set of $K$ actions, and  $\mathcal{K}^n \subset \mathcal{K}$ be the set of actions of the player $n$.
Let ${\bf y} \in [0,1]^K$ be a vector of bounded random variables, $y_{k}$ be the random variable denoting the reward of the action $k$ and $\mu_k$ be the mean reward of the action $k$.
Let $y^n_{k}$ be the random variable denoting the reward of the action $k$ chosen by the player $n$, and $\mu_k^n$ be its mean reward. 
Let $P({\bf y},n)$ be the joint distribution of rewards and {\it active} players.

\begin{algorithm}[!h]
  \caption{\sc Distributed Best Arm Identification Task}
  \label{DBA}
  {\bf Inputs:} $\epsilon \in [0,1)$, $\delta \in (0,1]$, $\forall n$ $\mathcal{K}^n=\mathcal{K}$\\
	{\bf Output:} an $\epsilon$-approximation of the best arm with probability $1-\delta$
\begin{algorithmic} [1]
		\REPEAT
			\STATE a player $n \sim P(n)$ is drawn
			\STATE a downward message is allowed
			\STATE update $\mathcal{K}^n$
			\STATE an action $k \in \mathcal{K}^n$ is played
			\STATE a reward $y^n_k \in [0,1]$ is received
			\STATE an upward message is allowed
		\UNTIL {$ \forall n$ and $ \forall k$ $\in \mathcal{K}^n$}, $\mu^n_k \geq \mu_{k^*}^n - \epsilon$
\end{algorithmic}
\end{algorithm}

\paragraph {Definition 1:}
an $\epsilon$-approximation of the best arm $k^*=\arg \max_{k \in \mathcal{K}} \mu_k$ is an arm $k \in \mathcal{K}$ such that $\mu_{k^*} \leq \mu_k + \epsilon$.

\paragraph {Definition 2:}
the sample complexity $t(A)$ is defined by the number of samples in $P(y)$ needed by the algorithm $A$ to obtain an $\epsilon$-approximation of the best arm with a probability $1-\delta$. 

\paragraph {Definition 3:}
the sample complexity $t_N(A)$ of the distributed algorithm $A$ on $N$ players is defined by the number of samples {\it per player} in $P(y)$ needed to obtain an $\epsilon$-approximation of the best arm with a probability $1-\delta$. 

\paragraph {}
In the following, the sample complexity of a rate optimal algorithm for the best arm identification problem is denoted $t^*$, and the sample complexity of a rate optimal distributed algorithm is denoted $t^*_N$.

\paragraph {Definition 4:}
for the best arm identification task, the speed-up factor $S(A)$ of the algorithm $A$ distributed on $N$ players with respect to an optimal algorithm run independently on each player 
is defined by:
\begin {displaymath}
S(A) = \max_{n \in \{1,...,N\}} \frac {T^*_n}{T_{N_\gamma}(A)} \text {,}
\end {displaymath}
where $T^*_n$ is the number of samples in $P({\bf y},n)$ needed to obtain on average $t^*$ draws of the player $n$, 
and $T_{N_\gamma}(A)$ is the number of samples in $P({\bf y},n)$ needed to obtain on average at least $t_{N_\gamma}(A)$ draws of each player included in $\mathcal {N}_\gamma$.

\paragraph {\bf Proposition 1:} {\it for the best arm identification task, the speed-up factor is greater or equal to the ratio between the sample complexity of an optimal algorithm run independently on each player and the one of the distributed algorithm $A$:}
\begin {displaymath}
S(A) \geq  \frac { t^*}{t_{N_\gamma}(A)}
\end {displaymath}

\begin {proof}
The number of times a player $n$ is drawn at time horizon $T$ is modeled by a binomial distribution of parameter T, $P(n)$.
At time step $T$ the mean number of draws of the player $n$ is $P(n)\cdot T$. 
This implies  that $\max_n T_n^* = \frac {t^*} {\min_n  P(n)}$, and $T_{N_\gamma}(A) = \frac {t_{N_\gamma}(A)} {\gamma} \leq \frac {t_{N_\gamma}(A)} {\min_n  P(n)}$.
\qed
\end {proof}

\paragraph {Assumption 1 (best arm identification task):} the mean reward of an action does not depend on the player: $\forall n$ $\in$ $ \{1,...,N\}$ and $\forall k$ $\in$ $ \{1,...,K\}$, $\mu^n_k=\mu_k$.
\paragraph {}
Assumption 1 is used to restrict the studied problem to the distribution of the best arm identification task. 
We discuss the extension of this distribution scheme to the {\it contextual bandit problem} in the future works.

\paragraph {Assumption 2 (binary code):} each transmitted message through the communication network is coded using a binary code\footnote {a {\it prefix code} such as a truncated binary code or a Huffman code (see \cite {CT06}) would be more efficient. To simplify the exposition of ideas, we have restricted the analysis to binary code.}.
For instance, when the synchronization server notifies to all players that the action $k=7$ is eliminated, it sends to all players the code $'111'$.

\paragraph{} 
In order to ease reading, in the following we will omit the algorithm $A$ in notations: $t_{N_\gamma}$ denotes the sample complexity of the distributed algorithm $A$ on $N_\gamma$ players, and $T_{N_\gamma}$ denotes the number of samples in $P({\bf y},n)$ needed to obtain on average at least $t_{N_\gamma}$ draws for each player included in $\mathcal {N}_\gamma$.
When assumptions 1 and 2 hold, Theorem~1 states a lower bound for this new problem. 

\paragraph {\bf Theorem 1:} 
{\it there exists a distribution $P({\bf y},n)$ such that,
any distributed algorithm on $N$ players needs to transmit at least $ 2N(K-1)\lceil \log_2 K \rceil$ bits  
to find with high probability an $\epsilon$-approximation of the best arm with an optimal speed-up factor in $\mathcal{O}(N)$.
}

\begin {proof}

Theorem~1 in \cite {MT04} states that there exists a distribution $P({\bf y})$ such that any algorithm needs to sample at least $\Omega\left( \frac {K}{\epsilon^2} \ln \frac{1}{\delta}\right)$ times to find with high probability an $\epsilon$-approximation of the best arm. 
As a consequence, the total number of draws of $N$ players needed by a distributed algorithm cannot be lesser than this lower bound.
Thus, there exists a distribution $P({\bf y})$ such that any distributed algorithm on $N$ players needs to sample at least $\Omega\left( \frac {K}{N .\epsilon^2} \ln \frac{1}{\delta}\right)$ times each player to find with high probability an $\epsilon$-approximation of the best arm.
When the distribution of players is uniform, we have $\mathcal {N} = \mathcal {N}_\gamma$, and hence for a distributed algorithm which is rate optimal in $\mathcal{O}(\frac {K}{N .\epsilon^2} \ln \frac{1}{\delta})$, we have:

\begin {displaymath}
S^* = \max_{n \in \{1,...,N\}} \frac {T^*_n}{T_N^*} = \frac {t^*} {t^*_N}=\mathcal {O}(N) \text {.}
\end {displaymath}

{\sc Median Elimination} \cite {EMM02} is a rate optimal algorithm for finding an $\epsilon$-approximation of the best arm. 
Thus, when the distribution of players is uniform, the speed-up factor of any distributed algorithm cannot be higher than $\mathcal{O}(N)$. \\

Let us assume that there exists a distributed algorithm that finds an $\epsilon$-approximation of the best arm with a speed-up factor $\mathcal{O}(N)$, 
and that transmits less than $2N.(K-1)\lceil \log_2 K \rceil$ bits. There are only three possibilities to achieve this goal:\\
(1) a player does not transmit information about an action to the server,\\
(2) or the server does not transmit information about an action to a player,\\
(3) or this algorithm transmits less than $\lceil \log_2 K \rceil$ bits for each action.

If a player does not transmit an information about an action to the server (condition $1$), then for this action the number of players is $N-1$. 
Thus, the speed-up factor $\mathcal{O}(N)$ cannot be reached.

If the server does not transmit information about an action to a player (condition $2$), 
then this player does not receive information about this action from the other players. 
As a consequence, this player cannot use information from other players to eliminate or to select this action, and in worst case the speed-up factor becomes $\mathcal{O}(1)$.

Thus, the number of sent messages cannot be less than $N.(K-1)$ upward messages plus $N(K-1)$ downward messages.
The minimum information that can be transmitted about an action is its index.
Using a binary code (see Assumption~2), the number of bits needed to transmit the index of an action cannot be less than $\lceil \log_2 K \rceil$ (condition $3$).
\qed
\end {proof}

\section {Distributed Median Elimination}

\subsection{Algorithm description}

\paragraph {}
Now, we can derive and analyze a simple and efficient algorithm to distribute the best arm identification task.
{\sc Distributed Median Elimination}  deals with three sets of actions:
\begin {enumerate}
\item $\mathcal{K}$ is the set of actions,
\item $\mathcal{K}^n$ is the set of remaining actions of the player $n$,
\item $\mathcal{K}^n_l$ is the set of actions that the player $n$ would not like to eliminate at local step $l$.
\end {enumerate}

{\sc Distributed Median Elimination} uses the most active players ($n \in \mathcal{N}_\gamma$) to eliminate suboptimal arms of local sets of actions $\mathcal{K}^n$ of all players.
When the algorithm stops, the players choose sequentially remaining actions from $\mathcal{K}^n$.
The sketch of the proposed algorithm (see Algorithm \ref {DSE}) is the following:
\begin {itemize}\item {\sc Median Elimination} algorithm with a (high) probability of failure $\eta = \delta^{\frac {2}{N_\gamma}}$ is run on each player, without the right of local elimination.
\item When a player would like to eliminate an action, the corresponding index of the action is sent to the synchronization server. 
\item When an half of the most active players would like to eliminate an action, the synchronization server eliminate the action with a (low) probability of failure $\eta^{N_\gamma/2}$
by sending the index of the eliminated action to each player.
\end {itemize}

\begin{algorithm}[!h]
  \caption{function MedianElimination($n$)}
  \label{ME}
\begin{algorithmic}[1]	
	\STATE Play sequentially an action $k \in \mathcal{K}^n$
	\STATE Receive $y^n_k$
	\STATE $t^n_k=t^n_k+1$, $\hat {\mu}^n_{k} = \frac {y^n_{k}}{t^n_k}+\frac{t^n_k-1}{t^n_k}\hat {\mu}^n_{k}$
	\IF {$\forall k  \in \mathcal{K}^n$ $t_k^n \geq 4/(\epsilon_l^n)^2 \ln(3K/\eta_l^n)$}
			\STATE Let $m_l^n$ be the median of $\hat {\mu}^n_{k}$ such that $k \in \mathcal{K}^n_l$
			\FOR {all $k \in \mathcal{K}^n_l$}
					\IF {$\left( \hat {\mu}^n_k < m_l^n \right)$}
						\STATE $\mathcal{K}^n_l=\mathcal{K}^n_l \setminus \{k\}$, $\lambda_k^n=1$, $\operatorname{Upward Message(k,n)}$
					\ENDIF
					\STATE $t^n_k=0$
			\ENDFOR
			\STATE $\epsilon^n_{l+1}=3/4 \cdot \epsilon^n_l$, $\eta^n_{l+1}=\eta^n_l/2$, $l=l+1$
	\ENDIF
\end{algorithmic}
\end{algorithm}
\begin{algorithm}[!h]
\caption {\sc Distributed Median Elimination}
\label {DSE}
\begin{algorithmic} [1]
	\STATE {\bf Inputs:} $0 < N_\gamma \leq N$, $\epsilon \in [0,1)$, $\delta \in (0,1]$, $\mathcal{K}$
	\STATE {\bf Output:} an $\epsilon$-approximation of the best arm in each set $\mathcal{K}^n$
	\STATE {\bf Synchronization server:} $\forall (k,n)$, $\lambda^n_k=0$ 
	\STATE {\bf Each player $n$:} $\eta = \delta^{\frac {2}{N_\gamma}}$, $\eta^n_1=\eta/2$ and $\epsilon^n_1=\epsilon/4$, $l=1$, $\mathcal{K}^n=\mathcal{K}$, $\mathcal{K}^n_l=\mathcal{K}$, $\forall k$ $t^n_k=0$, $\hat{\mu}^n_k=0$, $\lambda^n_k=0$
		\REPEAT
		\STATE a player $n \sim P(n)$ is drawn
		\STATE // Local process on player n
		\IF {$\left(\operatorname{Downward Message(k)} \text {\bf and } |\mathcal{K}^n| > 1\right)$}   
					\STATE $\mathcal{K}^n=\mathcal{K}^n \setminus \{k\}$, $\mathcal{K}^n_l=\mathcal{K}^n_l \setminus \{k\}$
		\ENDIF
		\STATE $\operatorname{MedianElimination}(n)$
		\STATE  // Process on synchronization server
			\IF {$\operatorname{Upward Message(k,n)}$}
				\STATE $\lambda^k_n=1$
				\IF {$\left( \sum_n \lambda^n_k \geq  N_\gamma/2\right)$} 
						\STATE $\operatorname{Downward Message(k)}$
				\ENDIF
			\ENDIF
			\UNTIL {$\left( \forall n, |\mathcal {K}^n_l| = 1 \right)$}
\end{algorithmic}
\end{algorithm}

\paragraph {Remark 1:}
{\sc Distributed Median Elimination} algorithm stops when all the most active players {\it would like} to eliminate all actions excepted their estimated best one ($\forall n, |\mathcal {K}^n_l|= 1$). This implies that each player can output several actions, and that the remaining actions are not necessary the same for each player.

\paragraph {}
The analysis is divided into four parts. The first part of the analysis insures that {\sc Distributed Median Elimination} algorithm finds an $\epsilon$-approximation of the optimal arm with high probability.
The second part states the communication cost in bits. The third part provides an upper bound of the number of pulls per player before stopping. 
The last part provides an upper bound of the number of samples in $P({\bf y},n)$ before stopping.

\subsection {Analysis of the algorithm output}

\paragraph {\bf Lemma 1:} {\it with a probability at least $1-\delta$, {\sc Distributed Median Elimination} finds an $\epsilon$-approximation of the optimal arm.}

\begin {proof} 
The proof uses similar arguments than those of Lemma 1 in \cite {EMM02}. The main difference is that here, for insuring that when the algorithm stops it remains an $\epsilon$-approximation of the best arm, we need to state that such near optimal arms cannot be eliminated with high probability until all sub-optimal arms have been eliminated.
Consider the event: 
\begin {displaymath}
E_1 = \{\exists k \in \mathcal{K}^n_l : \mu_k > \mu_{k^*} - \epsilon \text { and } \hat {\mu}^n_k < \mu_k - \epsilon^n_l/2\}.
\end {displaymath}
According to algorithm \ref {ME} line $4$, each arm is is sampled sufficiently such that:
\begin {displaymath}
P(\hat {\mu}^n_k < \mu_k - \epsilon^n_l/2) \leq \frac {\eta^n_l} {3K}.
\end {displaymath}
Using the union bound, we obtain that $P(E_1) \leq \eta^n_l/3$. \\

In case where $E_1$ does not hold, the probability that a suboptimal arm $k$ be empirically better than an $\epsilon$-approxi\-mation $k'$ of the best arm is:
\begin {displaymath}
P(\hat {\mu}^n_{k} \geq \hat {\mu}^n_{k'} | \neg E_1) \leq P(\hat {\mu}^n_{k} \geq \hat {\mu}^n_{k} + \epsilon^n_l/2 | \neg E_1)  \leq \frac {\eta^n_l} {3K}
\end {displaymath}

Let $B$ be the number of suboptimal arms, which are empirically better than an $\epsilon$-approxi\-mation of the best arm. Using Markov inequality, we have:
\begin {displaymath}
P( B \geq |\mathcal {K}^n_l|/2 | \neg E_1)  \leq \frac {2\mathds{E}[B]}{|\mathcal {K}^n_l|} \leq  \frac {2|\mathcal {K}^n_l|\eta^n_l}{3|\mathcal {K}^n_l|K} \leq \frac {2\eta^n_l} {3},
\end {displaymath}
where $\mathds{E}$ denotes the expectation with respect to the random variable ${\bf y}$.

\paragraph{}
As a consequence while it remains $|\mathcal{K}^n_l|/2$  suboptimal arms, an $\epsilon$-approxi\-mation of the best arm is not eliminated with a probability $1-\eta^n_l$. 
When the number of suboptimals arms is lesser than $|\mathcal{K}^n_l|/2$  lines $7-9$ of algorithm \ref {ME} insures that with a probability $1-\eta^n_l$ all the suboptimal arms are eliminated from $\mathcal{K}^n_l$, $\mathcal {K}^n_{l+1}$ is not empty, and $\mathcal {K}^n_{l+1}$ contains only $\epsilon$-approxi\-mations of the best arm. \\

Then using the union bound, the probability of failure is bounded by $\sum_{l=1}^{\log_2 K} \eta^n_l \leq \eta$.
By construction, the approximation error is reduced at each step such that $\sum_{l=1}^{\log_2 K} \epsilon^n_l \leq \epsilon$.
As a consequence when {\sc Distributed Median Elimination} stops, each set $\mathcal{K}^n_l$ contains an $\epsilon$-approximation of the best arm with a failure probability $\eta$. \\

{\sc Distributed Median Elimination} fails when it stops while $\exists n$ and $\exists k' \in \mathcal {K}^n$ such that $\mu_{k'} < \mu_{k*} - \epsilon$. 
This event could occur when $N_\gamma/2$ players would like to eliminate all $\epsilon$-approximations of the best arm, with a probability $\delta=\eta^{N_\gamma/2}$.

\qed
\end {proof}

\subsection {Analysis of the number of transmitted bits}

\paragraph {\bf Lemma 2:} {\it {\sc Distributed Median Elimination} stops transmitting $2N(K-1)\lceil \log_2 K \rceil$ bits.}

\begin {proof} 
Each action is sent to the server no more than once per player (see line $8$ of the algorithm \ref {ME}).
When the algorithm stops, the $N$ players have not sent the code of their estimated best action (see stopping condition line $19$ of the algorithm \ref {DSE}).
Thus the number of upward messages is $N(K-1)$.\\
Then, the fact that the synchronization server sends each suboptimal action only once insures that the number of downward messages is $N(K-1)$.\\
The optimal length of a binary code needed to code an alphabet of size $K$ is $\lceil \log_2 K \rceil$. \\
Thus, the total number of transmitted bits is $2N(K-1)\lceil \log_2 K \rceil$.
\qed
\end {proof}

\subsection {Analysis of the number of pulls per player}

\paragraph {\bf Lemma 3:} {\it {\sc Distributed Median Elimination} stops when each of the most actives player have been drawn at most}
\begin {equation*}
\mathcal{O}\left( \frac {K}{\epsilon^2 N_\gamma} \ln \frac{K}{\delta}\right) \text { \it times.}
\end {equation*}

\begin {proof}
The first steps of the proof are the same than those provided for {\sc Median Elimination} (see Lemma 2 in \cite {EMM02}). 
For the completeness of the analysis, we recall them here. From line 4 of Algorithm \ref {ME}, any player $n$ stops after:

\begin {equation*}
t_{N_\gamma} = \sum_{l=1}^{\log_2 K} \frac {4K^n_l}{(\epsilon^n_l)^2} \ln\frac {3K}{\eta^n_l} \text { pulls,} 
\end {equation*}
where $K^n_l$ is the number of actions at epoch $l$ of the player $n$. 
We have $\eta^n_l=\eta/2^l$, $\epsilon^n_l=(3/4)^{l-1}.\epsilon/4$, and $K^n_l = K/2^{l-1}$. Hence, we obtain:

\begin {equation}
\begin {split}
t_{N_\gamma} & \leq \frac {4}{\epsilon^2} \sum_{l=1}^{\log_2 K} \frac{K/2^{l-1}\ln(2^l\cdot 3K/\eta)} {\left( [3/4]^{l-1}\cdot \epsilon/4\right)^2}\\
 & \leq \frac {64K}{\epsilon^2} \sum_{l=1}^{\log_2 K} \left( \frac {8}{9} \right)^{l-1} \left(l \ln2+ \ln\frac {3K}{\eta}\right)\\
& \leq \frac {64K}{\epsilon^2} \ln \frac {K}{\eta} \sum_{l=1}^{\infty} \left( \frac{8}{9} \right)^{l-1} \left(l \cdot C_1+ C_2\right)\\
& \leq  \mathcal {O}\left(\frac {K}{\epsilon^2} \ln\frac {K}{\eta}\right)
\end {split}
\label {eq1}
\end {equation}

Replacing $\eta$ by $\delta^\frac{2}{N_\gamma}$ in inequality \ref {eq1},
we provide the upper bound of the number of pulls per player.
\qed
\end {proof}

Theorem~2 states that when $N_\gamma=N$ the speed-up factor of {\sc Distributed Median Elimination} is at least in $\mathcal {O}(N/(1+ \ln K))$ with respect to an optimal algorithm such as {\sc Median Elimination} \cite {EMM02} run on each player, while its communication cost is at most $2N(K-1)\lceil \log_2 K \rceil$ bits. 
Theorem~1 and Theorem~2 show that {\sc Distributed Median Elimination} is optimal in term of number of transmitted bits and near optimal in term of speed-up factor.

\paragraph {\bf Theorem 2:} {\it when $N=N_\gamma$ with a probability at least $1-\delta$, {\sc Distributed Median Elimination} finds with high probability an $\epsilon$-approximation of the best arm, 
transmitting $2N(K-1)\lceil \log_2 K \rceil$ bits, and obtains a speed-up factor at least in $\mathcal{O}\left(N/(1+ \ln K)\right)$.} 

\begin {proof}
Using Lemma 1, 2, and 3, we state that {\sc Distributed Median Elimination} finds  with high probability an $\epsilon$-approximation of the best arm, 
transmitting $2N(K-1)\lceil \log_2 K \rceil$ bits through the communication network, and using no more than\\
$\mathcal{O}\left( \frac {K}{\epsilon^2 N} \ln \frac{K}{\delta}\right)$ pulls per player.
{\sc Median Elimination} is an optimal algorithm for finding an $\epsilon$-approxima\-tion of the best arm: its sample complexity reaches the lower bound 
in $\Omega\left( \frac {K}{\epsilon^2} \ln \frac{1}{\delta}\right)$ pulls.
Thus using Proposition 1, the speed-up factor of {\sc Distributed Median Elimination} is:
\begin {equation*}
\mathcal{O}\left( N\frac {\ln 1/\delta}{\ln K/\delta}\right) \geq \mathcal{O}\left( \frac {N}{1+\ln K }\right)
\end {equation*}
\qed
\end {proof}

\subsection {Analysis of the number of draws of players}

\paragraph {}

The analysis of the number of pulls per player allows to state a near optimal speed-up factor in $\mathcal{O}(N/(1+\ln K))$.
Now, we focus on the number of draws of players (i.e. the time step) needed to insure with high probability that all players find an $\epsilon$-approximation of the best arm.
First, we consider the case where the true value of $N_\gamma$ is known.
This requires some knowledge of the distribution of players, which is realistic in many applications. 
For instance, in the case of Radio Access Network, the load of each cell or server is known, and hence the probability of each player is known.
Theorem~3 provides an upper bound of the number of draws of players needed to find 
an $\epsilon$-approximation of the best arm with high probability, when $N_\gamma$ is known. 
In the next section we consider the case where $N_\gamma$ is unknown.

\paragraph {\bf Theorem 3:} {\it with a probability at least $1-\delta$, {\sc Distributed Median Elimination} finds an $\epsilon$-approxima\-tion of the best arm, 
transmitting $2N(K-1)\lceil \log_2 K \rceil$ bits through the communication network and using at most}

\begin {equation*}
 \mathcal{O}   \left( 
\left(\frac {K}{\epsilon^2 \gamma N_\gamma} + \sqrt{ \frac {K}{\epsilon^2N_\gamma}  }\right) \ln \frac{KN_\gamma}{\delta}  
\right)   
\text{\it draws of players,}
\end {equation*}
{\it where } $\gamma \in (0,1]$ and $N_\gamma=|\{n \leq N, P(n \in \mathcal{N}_\gamma) > \gamma \}|$.

\paragraph{}
\begin {proof}
Consider the event $n \notin \mathcal {N}_\gamma$. We have $ P(n \notin {N}_\gamma) = 1-\gamma$.
Let $f$ be the number of times where a player has not been drawn at time step $T$.
$f$  follows a negative binomial distribution with parameters $t$,$1-\gamma$.
By definition of the negative binomial distribution, We have:

\begin {displaymath}
\mathds{E}[f]=\frac {(1-\gamma) t}{\gamma}, 
\end {displaymath}
where $\mathds{E}$ denotes the expectation with respect to the random varible $n$.
Using the Hoeffding's inequality, we have:

\begin {displaymath}
P \left( f - \frac {(1-\gamma) t}{\gamma}  \geq \epsilon.t \right) \leq \exp(-2\epsilon^2\cdot t^2) = \alpha
\end {displaymath}
\begin {displaymath}
\Leftrightarrow P \left( f   \geq \frac {(1-\gamma) t}{\gamma}+\sqrt{\frac{t}{2}\ln \frac{1}{\alpha}}  \right) \leq \alpha
\end {displaymath}

The number of draws $T$ is the sum of $t$, the number of draws of a player, and $f$, the draws which do not contain this player. 
Hence, setting $\alpha=\frac{\delta}{2N_\gamma}$, using Lemma 1 with a failure probability $\frac{\delta}{2N_\gamma}$, and then using
the union bound, the following inequality is true with a probability $1-\delta$:

\begin {displaymath}
\begin {split}
T_{N_\gamma} & \leq   t_{N_\gamma} + \frac {(1-\gamma) t_{N_\gamma}}{\gamma} +\sqrt{\frac{t_{N_\gamma}}{2}\ln \frac{2N_\gamma}{\delta}}\\
 & \leq  \frac {t_{N_\gamma}}{\gamma}  +\sqrt{\frac{t_{N_\gamma}}{2}\ln \frac{2N_\gamma}{\delta}}
 \end {split}
\end {displaymath}

Using Lemma 3, we have:

\begin {equation*}
\begin {split}
T_{N_\gamma} & \leq     
\frac {K}{\epsilon^2 \gamma N_\gamma} \ln \frac{2KN_\gamma}{\delta}
+ \sqrt{ \frac {K}{2\epsilon^2N_\gamma}\ln \frac{2KN_\gamma}{\delta} \ln \frac{2N_\gamma}{\delta} } \\
&  \leq \mathcal{O}     \left( 
\left(\frac {K}{\epsilon^2 \gamma N_\gamma} + \sqrt{ \frac {K}{\epsilon^2N_\gamma}  }\right) \ln \frac{KN_\gamma}{\delta}  
\right)
 \end {split}
\end {equation*}

Then using Lemma 2, we conclude the proof.
\qed
\end {proof}

\section {Extended Distributed Median Elimination}
\subsection {Algorithm description}
\paragraph{}
Notice that if $N_\gamma$ is not gracefully set, the stopping time of {\sc Distributed Median Elimination} is not controlled.
The algorithm could not stop if $\mathcal {N}_\gamma$ contains a player with a zero probability, or could stop after a lot of time steps, if $\mathcal {N}_\gamma$ contains an unlikely player.
Moreover, {\sc Distributed Median Elimination} is designed to transmit an optimal number of bits. This first algorithm cannot handle the trade-off between the number of time steps, where all players have selected an $\epsilon$-approximation of the best arm, and the number of transmitted bits.
To overcome these limitations, we propose a straightforward extension of the proposed algorithm, which consists in playing in parallel $M$ instances of {\sc Distributed Median Elimination} with equally spread values of $N_\gamma$ (see Algorithm \ref {DSE2}). 

\begin{algorithm}[!h]
\caption {\sc Extended Distributed Median Elimination}
\label {DSE2}
\begin{algorithmic} [1]
	\STATE {\bf Inputs:} $\forall i$ $\in \{1,...,M\}$ $N_{\gamma_i}=N.i/M$, $\epsilon \in [0,1)$, $\delta \in (0,1]$, $\mathcal{K}$
	\STATE {\bf Output:} an $\epsilon$-approximation of the best arm in each set $\mathcal{K}^n$
	\STATE {\bf Synchronization server:} $\forall (k,n,i)$, $\lambda^{n_i}_k=0$ 
	\STATE {\bf Each player $n$:}	$\forall i$ $\eta_i = (\delta/M)^{\frac {2}{N_{\gamma_i}}}$, 
 $\eta^{n_i}_1=\eta_i/2$ and $\epsilon^{n_i}_1=\epsilon/4$, $l=1$, $\mathcal{K}^{n}=\mathcal{K}$, $\mathcal{K}^{n_i}_l=\mathcal{K}$, $\forall (k,i)$ $t^{n_i}_k=0$, $\hat{\mu}^{n_i}_k=0$, $\lambda^{n_i}_k=0$
	\REPEAT
		\STATE a player $n \sim P(n)$ is drawn
		\STATE // Local process on player n
		\FOR {$i \in \{1,...,M\}$}
			\IF { $\operatorname{Downward Message(k)} \text {\bf and } |\mathcal{K}^{n}| > 1$}   
					\STATE $\mathcal{K}^{n}=\mathcal{K}^{n} \setminus \{k\}$, $\mathcal{K}_l^{n_i}=\mathcal{K}_l^{n_i} \setminus \{k\}$
			\ENDIF
			\STATE $\operatorname{MedianElimination}(n_i)$
		\ENDFOR
		\STATE // Process on synchronization server
		\FOR {each $\operatorname{Upward Message(k,n_i)}$}
			\STATE $\lambda^k_{n_i}=1$
		\ENDFOR
		\IF {$\left(\exists i \text { such that }\sum_{n_i} \lambda^{n_i}_k \geq  N_{\gamma_i}/2\right)$}
						\STATE $\operatorname{Downward Message(k)}$
		\ENDIF
	\UNTIL {$\left(  \exists i \text { such that } \forall n_i, |\mathcal {K}^{n_i}_l| = 1 \right) $}
\end{algorithmic}
\end{algorithm}

\subsection {Analysis}

\paragraph {\bf Theorem 4:} {\it with a probability at least $1-\delta$,  {\sc Extended Distributed Median Elimination} finds an $\epsilon$-approximation of the best arm, transmitting $2NM(K-1)\lceil \log_2 K \rceil$ bits through the communication network and using at most}

\begin {equation*}
 \mathcal{O}     \left(\min_{i \in \{1,...,M\}} \left[  
\left(\frac {K}{\epsilon^2 \gamma_i N_{\gamma_i}} + \sqrt{ \frac {K}{\epsilon^2N_{\gamma_i}}  }\right)  \ln \frac{KMN_{\gamma_i}}{\delta} 
\right]
\right)
\end {equation*}
{\it draws of players, where } $0 < \gamma_{M} \leq ... \leq \gamma_i \leq... \leq \gamma_1 \leq 1$, and $N_{\gamma_i}=N.i/M$.

\begin {proof}
The proof of Theorem~4 straightforwardly comes from Theorem~3. 
Theorem~3 holds for each instance with a probability $1-\delta/M$. Then using the union bound, Theorem~3 holds for all instances with a probability $1-\delta$.
The communication cost is the sum of communications of each instance, and the number of time steps needed to find a near optimal arm is the minimum of all instances.
\qed

\end {proof}

{\sc Extended Distributed Median Elimination}\\
handles the trade-off between the time step where all players have chosen a near optimal arm and the number of transmitted bits (see Theorem 4).
The communication cost increases linearly with $M$, while the needed number of draws decreases with $M$.
Moreover, one can insure that the algorithm stops by setting an instance where $N_\gamma=1$, and that the algorithm has a speed-up factor in $\mathcal{O}(N/(1+ \ln K))$ in the worst case (i.e. when the distribution of players is uniform) by setting another instance of the algorithm where $N_\gamma=N$.
The cost of this good behavior is the factor $M$ in the communication cost.

\section {Experiments}

\subsection {Experimental setting}

\paragraph {}
In this section we provide and discuss some experiments done with a simulated environment. 
To illustrate and complete the analysis of the proposed algorithm, we compare {\sc Distributed Median Elimination} on regret minimization problems using three baselines:
{\sc Median Elimination} \cite {EMM02} played independently on each player and {\sc Median Elimination} with an unlimited communication cost $(64T\times\log_2K)$ illustrate the interest and the limits of the distribution approach, and {\sc UCB} \cite {ABF02} with an unlimited communication cost is used as a benchmark for the two regret minimization problems.
In order to finely capture the difference of performances between distributed and non distributed algorithm, we plot the estimated pseudo-regret over time:
\begin {displaymath}
 R(T) = T \cdot \mu_{k^*} - \sum_{t=0}^T \bar{y}_{k_t}^{n_t} \text { , }
\end {displaymath}
where $k_t$ is the action chosen by the player $n_t$ drawn at time $t$, and $\bar{y}_{k_t}^{n_t}$ is the estimated reward over $100$ trials.

\paragraph {Problem 1.}
There are $10$ arms. The optimal arm has a mean reward $\mu_0=0.7$, the second one $\mu_1=0.5$, the third one $\mu_2=0.3$, and the others have a mean reward of $0.1$. 
The problem 1, where a few number of arms have high mean rewards and the others have low mean rewards, is easy for an explore-then-commit strategy such as {\sc Distributed Median Elimination} 
and for a fully-sequential approach such as {\sc UCB}.

\paragraph {Problem 2.}
There are $10$ arms. The optimal arm has a mean reward $\mu_0=0.3$, the second one $\mu_1=0.2$, and the other have a mean reward of $0.1$. 
With regard to this more difficult problem, where the gap between arms is tighter, explore-then-commit and fully-sequential algorithms will need more steps to play frequently the best arm.
In contrast, {\sc Median Elimination} is a fixed-design approach: whatever the problem, it spends the same number of steps in exploration.

\paragraph{}
For highlighting the interest of using {\sc Distributed Median Elimination} for the proposed problem setting (see Figure \ref {fig1} and Algorithm \ref {DBA}) and for ensuring a fair comparison between algorithms, two distribution of players are tested.

\paragraph {Uniform distribution.}
Each player has a probability equal to $1/N$. 
In this case, the knowledge of the distribution of players does not provide any particular benefit for {\sc Distributed Median Elimination}:  $N_\gamma$ is set to $N$, which is known by all blind algorithms. This case corresponds to the worst case for {\sc Distributed Median Elimination}.

\paragraph {$20 \%$ of players generates $80 \%$ of events.}
The players are part in two groups of sizes $N_\gamma$ and $N-N_\gamma$.
When a player is drawn, a uniform random variable $x \in [0,1]$ is drawn. If $x < 0.8$  the player belongs from the first group, and else from the second one.
In this case, the knowledge of the distribution provides a useful information for setting $N_\gamma=0.2\times N$.
This knowledge, which corresponds to the number of most active players, is available for cells in a Radio Access Network.

\paragraph {}
For all the experiments, $\epsilon$ is set to $0.5$, $\delta$ is set to $0.05$, and the time horizon is $10^6$. All the regret curves are averaged over $100$ trials.

\subsection {Discussion}

\paragraph {}
We notice that the number of transmitted bits is zero for {\sc Median Elimination} played independently on each player, 
$4736$ for {\sc Distributed Median Elimination} run on $64$ players, and $2.56 \times 10^8$ for {\sc Median Elimination} and {\sc UCB} with an unlimited communication cost. 
In comparison to algorithms with unlimited communication cost, {\sc Distributed Median Elimination} needs $10^6$ times less bits to process a million of decisions: the communication cost of {\sc Distributed Median Elimination} does not depend on the time horizon (see Lemma~2).

\begin{figure}[htp]
  \centering
  \subfigure[The number of players versus the regret at time horizon $10^6$]{\includegraphics[width=8.3 cm]{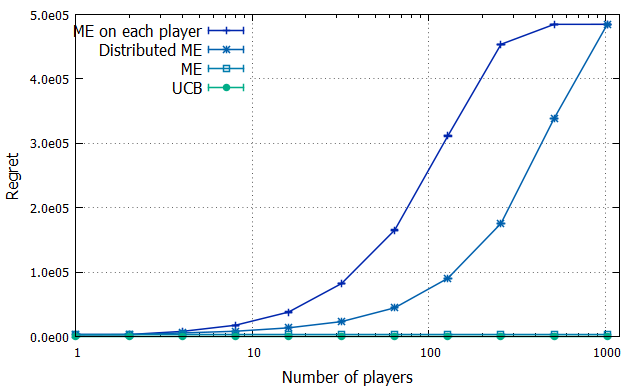}}
  \hspace{1pt}
  \subfigure[The time horizon versus the regret for $64$ players]{\includegraphics[width=8.3 cm]{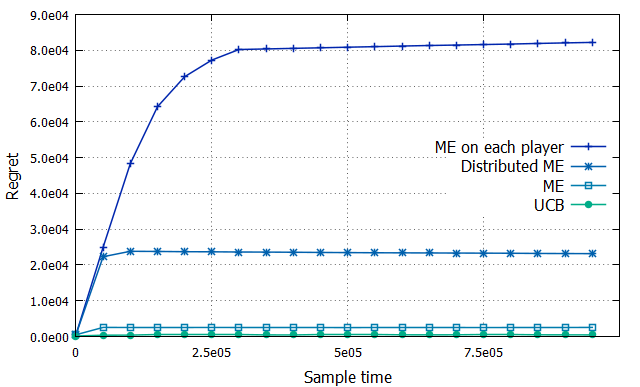}}
  \caption{Problem 1 - Uniform distribution of players}
  \label{Toy1}
\end{figure}

\begin{figure}[htp]
  \centering
  \subfigure[The number of players versus the regret at time horizon $10^6$]{\includegraphics[width=8.3 cm]{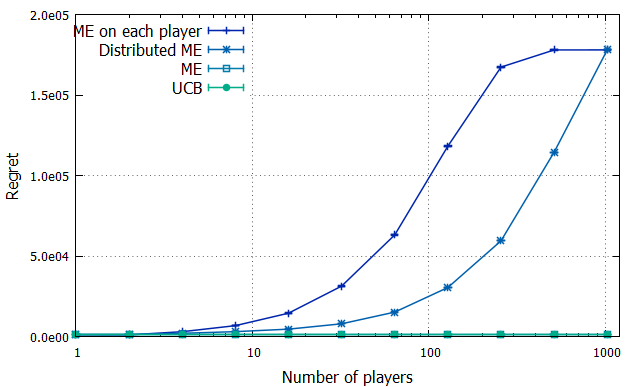}}
  \hspace{1pt}
  \subfigure[The time horizon versus the regret for $64$ players]{\includegraphics[width=8.3 cm]{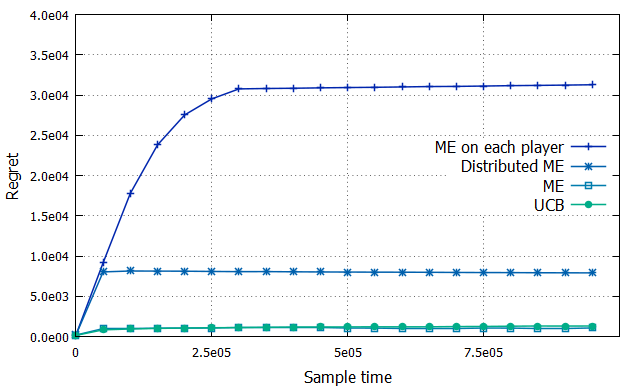}}
  \caption{Problem 2 - Uniform distribution of players}
  \label{Toy2}
\end{figure}

\paragraph {}
The number of players versus the regret at time horizon $10^6$ is plotted for the two problems when the distribution of players is uniform (see Figures \ref {Toy1}a and \ref {Toy2}a). 
Firstly, we observe that whatever the number of players {\sc Distributed Median Elimination} is outperformed by {\sc Median Elimination} with an unlimited communication cost. 
For $1024$ players, the expected number of events per player is lower than one thousand: {\sc Distributed Median Elimination} and {\sc Median Elimination} performed on each player does not end the first elimination epoch. Secondly, for less than $1024$ players, {\sc Distributed Median Elimination} clearly outperforms {\sc Median Elimination} with a zero communication cost.

\begin{figure}[htp]
  \centering
  \subfigure[The number of players versus the regret at time horizon $10^6$]{\includegraphics[width=8.3 cm]{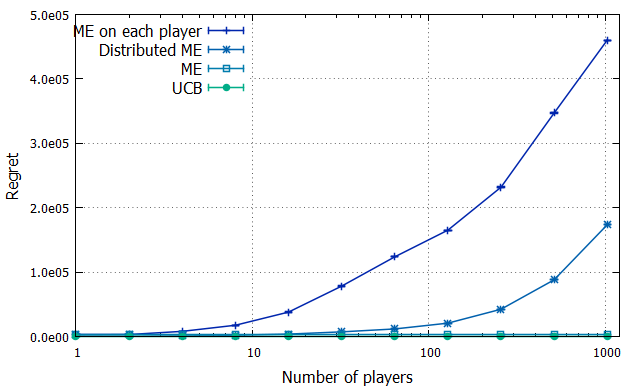}}
  \hspace{1pt}
  \subfigure[The time horizon versus the regret for $64$ players]{\includegraphics[width=8.3 cm]{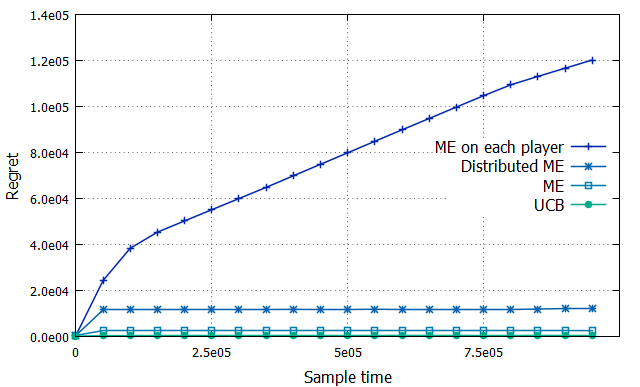}}
  \caption{Problem 1 - $20 \%$ of players generates $80 \%$ of events}
  \label{Toy3}
\end{figure}

\paragraph {}
The regret versus the time step is plotted for the two problems using $64$ players which are uniformely distributed (see Figures \ref {Toy1}b and \ref {Toy2}b).
For the first problem (see Figure \ref {Toy1}b), where the gap is large, {\sc UCB} with an unlimited communication cost benefits from its fully-sequential approach: it outperforms clearly {\sc Median}\\ {\sc Elimination}. The second problem (see Figure \ref {Toy2}b) is more difficult since the gap is tighter. As a consequence, the difference in perfomances between {\sc UCB} and {\sc Median Elimination} is small.
{\sc Distributed Median}\\ {\sc Elimination} significantly outperforms {\sc Median Elimination} with zero communication cost on both problems.

\begin{figure}[htp]
  \centering
  \subfigure[The number of players versus the regret at time horizon $10^6$]{\includegraphics[width=8.3 cm]{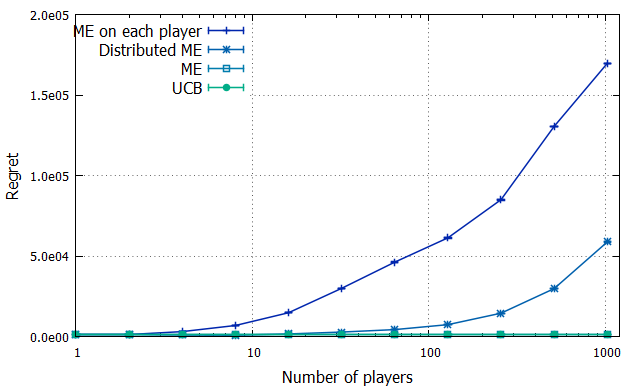}}
  \hspace{1pt}
  \subfigure[The time horizon versus the regret for $64$ players]{\includegraphics[width=8.3 cm]{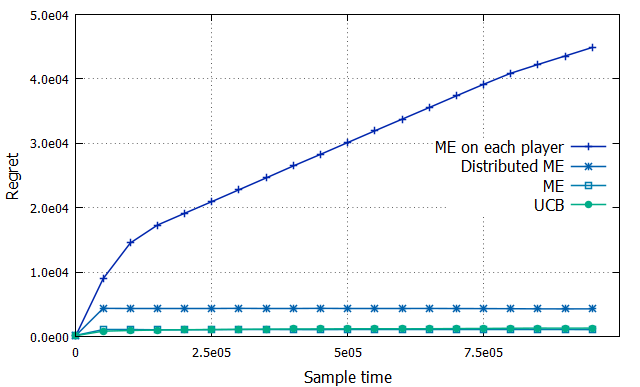}}
  \caption{Problem 2 - $20 \%$ of players generates $80 \%$ of events}
  \label{Toy4}
\end{figure}

\paragraph {}
When the distribution of players is not uniform, we observe that the gap in performances between {\sc Median Elimination} with an unlimited communication cost and {\sc Distributed Median Elimination} is reduced (see Figures \ref {Toy3}a and \ref {Toy4}a). In comparison to {\sc Median Elimination} played on each player, {\sc Distributed Median Elimination} exhibits a good behavior: when the most active players have found an $\epsilon$-approximation of the best arm, the sharing of information allows to eliminate the suboptimal arms for infrequent players which are numerous (see Figure \ref {Toy3}b and \ref {Toy4}b). As a consequence, the gap in performances between {\sc Distributed Median Elimination} and {\sc Median Elimination} played on a single player is increased.
\begin{figure}[htp]
  \centering
  \subfigure[Problem 1]{\includegraphics[width=8.3 cm]{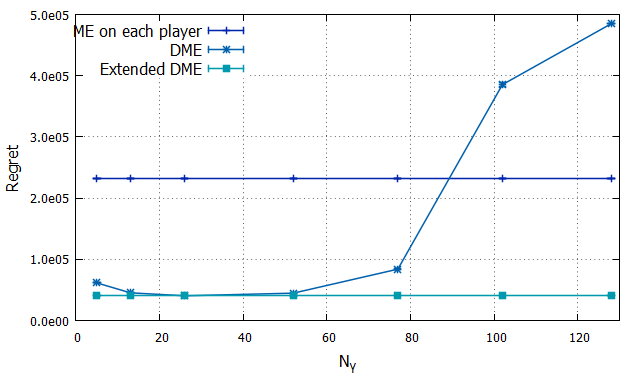}}
  \hspace{1pt}
  \subfigure[Problem 2]{\includegraphics[width=8.3 cm]{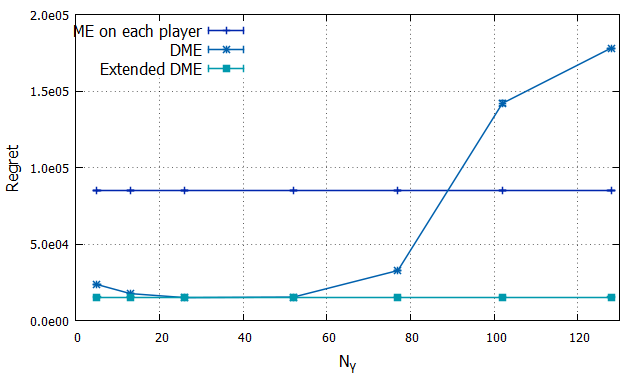}}
  \caption{The parameter $N_\gamma$ versus the regret when $20 \%$ of players generates $80 \%$ of events}
  \label{Toy5}
\end{figure}

\paragraph{}
To illustrate the interest of {\sc Extended Distributed Median Elimination} when the knowledge of the distribution of players is not available, 
the value of the parameter $N_\gamma$ versus the regret at the time horizon is plotted (see Figure \ref {Toy5}) for the two problems with $256$ players.
When $20 \%$ of players generates $80 \%$ of events, {\sc Distributed Median Elimination} outperforms {\sc Median Elimination} run on each player for a wide range of values of the parameter $N_\gamma$. However, when $N_\gamma$ is overestimated, the speed-up factor with respect to {\sc Median Elimination} run on each player can be lesser than one. Without the knowledge of the true value of the parameter, by selecting the best instance {\sc Extended Distributed Median Elimination} obtains the result of the best instance of {\sc Distributed Median Elimination} and significantly outperforms {\sc Median Elimination} run on each player. The communication cost becomes $75776$ bits instead $9472$ bits, when $8$ instances run in parallel.

\section {Conclusion an future works}

\paragraph {}
In order to distribute the best identification task as close as possible to the user's devices, we have proposed a new problem setting, where the players are drawn from a distribution. This architecture guarantees privacy to the users since no data are stored and the only thing that can be observed by an adversary through the core network is aggregated information over users.
When the distribution of players is known, we provided and analyzed a first algorithm for this problem: {\sc Distributed Median Elimination}.
We have showed that its communication cost is optimal, while its speed-up factor in $\mathcal{O}(N/(1+\ln K))$ is near optimal.
Then, we have proposed {\sc Extended Distributed Median Elimination}, which handles the trade-off between the communication cost and the speed-up factor.
In four illustrative experiments, we have compared the proposed algorithm with three baselines: {\sc Median Elimination} with zero and unlimited communication costs, and {\sc UCB} with an unlimited communication cost.
According to the theoretical analysis, {\sc Distributed Median Elimination} clearly outperforms {\sc Median Elimination} with a zero communication cost.
Finally, this distribution approach provides a speed-up factor linear in term of number of Mobile Edge Computing application server, facilitates privacy by processing data close to the end-user, and its communication cost, which does not depend on the time horizon, allows to control the load of the telecommunication network while deploying a lot of decision making applications on the edge of the Radio Access Network.

\paragraph {}
These results are obtained when Assumption $1$ holds: the mean reward of actions does not depend on the player.
Future works will extend this distributed approach to the case where Assumption $1$ does not hold, and in particular for the contextual bandit problem.
Indeed, {\sc Distributed Median Elimination} is a basic block, which can be extended to the selection of variables to build a distributed decision stump and then a distributed version of {\sc Bandit Forest} \cite {FAUC16}.



\end{document}